\documentclass[a4paper,fleqn]{cas-dc}

\usepackage[numbers]{natbib}
\usepackage{adjustbox}

\usepackage{algorithm}
\usepackage{algpseudocode}
\usepackage[normalem]{ulem}
\definecolor{ourscolor}{HTML}{DCEBFF}

\def\tsc#1{\csdef{#1}{\textsc{\lowercase{#1}}\xspace}}
\tsc{WGM}
\tsc{QE}
\begin{document}
\let\printorcid\relax  % Remove ORCID(s) footer
\let\WriteBookmarks\relax
\def\floatpagepagefraction{1}
\def\textpagefraction{.001}

\shorttitle{}

\shortauthors{Jia et al.}

% Main title
\title[mode = title]{PI-MOT: Prior-Informed Interaction Regularization for Multi-Object Tracking}

% ==================== Authors ====================

% Author 1
\author[1,2]{Shukun Jia}
% \ead{Email}
% \credit{Conceptualization, Methodology, Software, Writing - original draft}

% Author 2
\author[3]{Shiyu Hu}
% \ead{Email}

% Author 3
\author[1,2]{Yipei Wang}
% \ead{Email}

% Author 4
\author[1,2]{Ximeng Cheng}
% \ead{Email}

% Author 5
\author[4]{Xin Zhao}
% \ead{Email}

% Author 6 (Corresponding author)
\author[1,2]{Xiaobo Lu}
\cormark[1]
\ead{xblu2013@126.com}

\cortext[1]{Corresponding author.}

% ==================== Affiliations ====================

\affiliation[1]{
    organization={School of Automation},
    addressline={Southeast University},
    city={Nanjing},
    postcode={210096},
    country={China}
}

\affiliation[2]{
    organization={Key Laboratory of Measurement and Control of Complex Systems of Engineering},
    addressline={Ministry of Education},
    city={Nanjing},
    postcode={210096},
    country={China}
}

\affiliation[3]{
    organization={School of Physical \& Mathematical Sciences},
    addressline={Nanyang Technological University},
    postcode={639798},
    country={Singapore}
}

\affiliation[4]{
    organization={School of Computer and Communication Engineering},
    addressline={University of Science and Technology Beijing},
    city={Beijing},
    postcode={100083},
    country={China}
}

\begin{abstract}
Tracking-by-query (TBQ) has emerged as an important end-to-end paradigm for multi-object tracking (MOT), unifying object localization and identity association through learnable query interaction. However, conventional affinity-driven interaction leaves task-specific regularities largely implicit, placing a greater burden on data-driven learning and increasing the risk of unreliable association under ambiguous observations. To address this issue, we reformulate query interaction as prior-informed information routing and propose PI-MOT, which encodes tracking regularities as soft, differentiable biases within the interaction process. PI-MOT derives reciprocal support between track and detection queries from spatially modulated affinity and uses this support to regulate both observation aggregation and reliability-aware temporal state updates.
Extensive experiments demonstrate strong performance on three challenging benchmarks, with HOTA scores of 70.7 on DanceTrack, 77.0 on SportsMOT, and 62.5 on MOT20. Further analyses on DanceTrack show that PI-MOT achieves a favorable balance between tracking performance and learning efficiency, converging faster while retaining strong performance with substantially less training data. The source code will be made publicly available at
\url{https://github.com/FaithFlow/PI-MOT}.
\end{abstract}


\begin{highlights}
\item Tracking priors regularize query interaction through soft, differentiable biases.
\item Spatially informed reciprocal support reduces ambiguous query competition.
\item Interaction confidence guides track-state and historical memory updates.
\item PI-MOT achieves 70.7, 77.0 and 62.5 HOTA on DanceTrack, SportsMOT and MOT20.
\item With 25\% of DanceTrack training data, PI-MOT retains over 95\% of full-data HOTA.

\end{highlights}

%\nocite{*}

% Keywords

\maketitle

\section{Introduction}
\label{sec:intro}
Multi-object tracking (MOT) aims to continuously localize objects and maintain their identities across video sequences, serving as fundamental support in a wide range of vision systems such as autonomous driving and video analytics \cite{hu2023_uniad, kao2026sritrack}. Despite significant advances, reliable identity association remains challenging under frequent crossovers, severe occlusions, and dense interactions. 

Traditional tracking-by-detection (TBD) methods typically rely on heuristic association strategies that explicitly incorporate motion and spatial constraints \cite{zhang2022bytetrack,cao2023observation,yang2024hybrid}. Their association mechanisms are interpretable, but largely governed by predefined rules. In contrast, tracking-by-query (TBQ) frameworks enable end-to-end association through learnable query interaction \cite{zeng2022motr,zhu2022looking,meinhardt2022trackformer}. Specifically, detection queries represent current observations, while track queries maintain identity information across frames \cite{zhang2023motrv2,yan2025comot,gao2025multiple,jia2026tracking}. Their interaction determines how observations are integrated into track representations, thereby influencing the evolution of identity states over time.

\begin{figure*}[pos=t]
    \centering
    \includegraphics[width=0.88\textwidth]{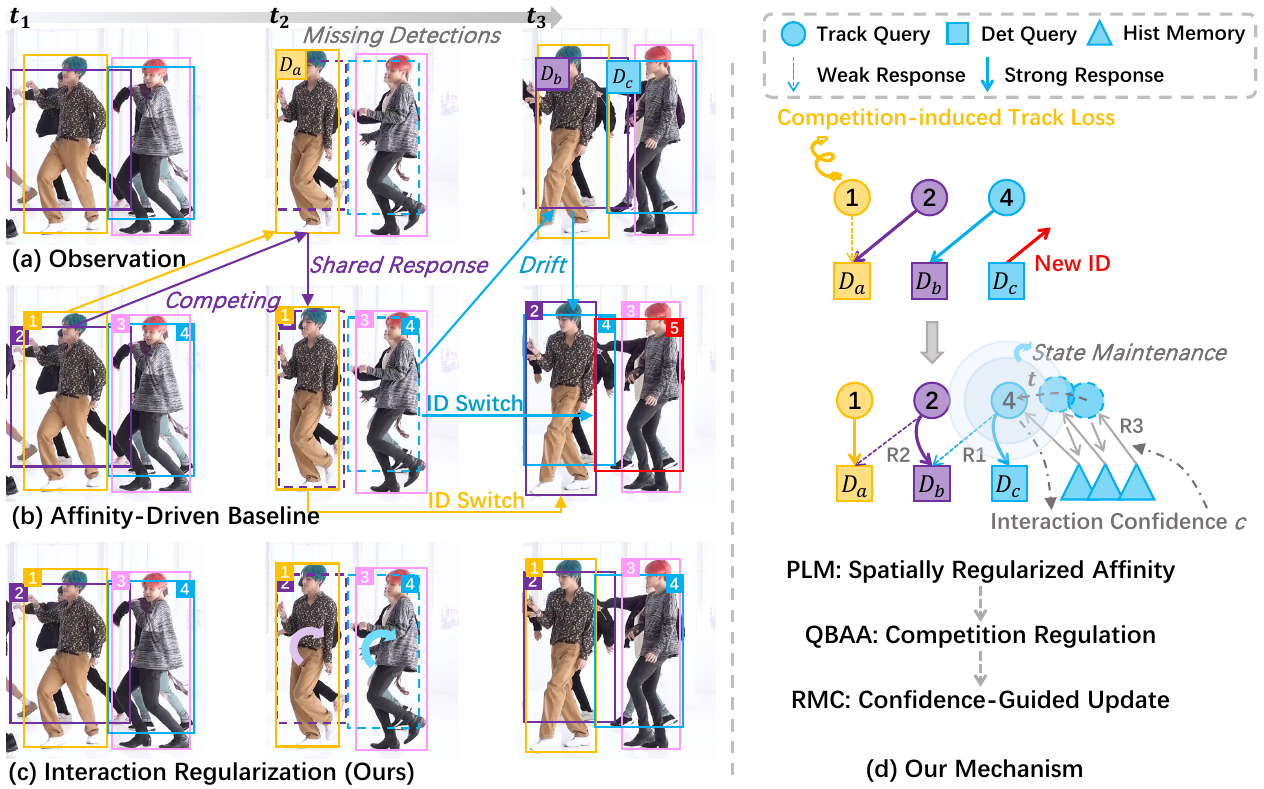}
    \caption{
    \textbf{Illustration of ambiguous identity association and structurally regularized query interaction.} (a) A representative challenge in MOT, where severe occlusion can lead to missing detections, reducing the available evidence for identity association. (b) Under such conditions, the affinity-driven baseline exhibits competing associations and identity drift, leading to tracking errors such as ID switches, track loss, and erroneous new track initialization. (c) In contrast, PI-MOT maintains consistent track identities under the same challenging observations. (d) Conceptual illustration of the proposed mechanism at $t_2$ and $t_3$. Spatially modulated affinities from PLM (R1) are used by QBAA to assess reciprocal track–detection support (R2). RMC derives interaction confidence from this support to regulate track-state and memory updates (R3), increasing reliance on historical memory when current support is weak.
    }
    \label{fig:concept}
    \vspace{-10pt}
\end{figure*}

However, learnable query interaction can exhibit fragile behavior under ambiguous tracking conditions. As illustrated in Fig.~\ref{fig:concept}(a–b), severe occlusion can lead to missing detections, causing competing responses among track queries and potentially inducing identity drift through unreliable information integration. As the affected
track representations are propagated across frames, this interference extends to subsequent identity association.

More generally, reliable identity association involves integrating current detections while preserving identities established in previous frames. Temporal continuity in object locations provides a basic cue for relating a detection to an existing track \cite{zhang2022bytetrack,yang2024hybrid,shim2025focusing}. Under ambiguity, a detection may elicit strong responses from several tracks; comparing these possible matches helps distinguish correspondences with stronger support from ambiguous alternatives. The overall support from current observations, in turn, informs how much a track should rely on them. When this support is insufficient, information retained from earlier moments can help maintain identity continuity. Selecting current evidence and integrating it over time therefore involve interdependent decisions about which observations support a track and how much to rely on them. Conventional affinity-driven interaction leaves these dependencies largely implicit. Learning them from tracking supervision may require additional training evidence to establish reliable interactions under ambiguity.

Motivated by this task structure, we propose PI-MOT, a prior-informed framework that embeds tracking inductive biases into learnable query interaction to regularize information routing. We use track–detection support to determine both the relative contributions of current observations and their influence on track-state and memory updates. Specifically, Proximity-Guided Logit Modulation (PLM) incorporates spatial consistency into learned affinity, and Quasi-Bijective Attentional Alignment (QBAA) evaluates reciprocal support from the resulting affinities to alleviate ambiguous competition. Reliability-Guided Memory Compensation (RMC) derives interaction confidence from this support to balance current observations and historical memory, increasing reliance on memory when current support is weak and regulating memory updates accordingly (Fig.~\ref{fig:concept}(c–d)).

Extensive experiments demonstrate the effectiveness of PI-MOT across challenging MOT benchmarks, achieving HOTA scores of 70.7 on DanceTrack, 77.0 on SportsMOT, and 62.5 on MOT20. Beyond overall tracking performance, PI-MOT exhibits more reliable interaction behavior, faster convergence, and improved data efficiency. Notably, using only 25\% of the DanceTrack training data, PI-MOT achieves 67.6 HOTA on the test benchmark, retaining over 95\% of its full-data performance.

The main contributions are summarized as follows:

\begin{itemize}

\item We reformulate affinity-driven query interaction as structurally regularized information routing to promote robust association and data-efficient interaction learning.

\item We instantiate this formulation in PI-MOT, an end-to-end tracking framework built around lightweight, fully differentiable interaction regularization. PLM and QBAA establish spatially informed reciprocal track–detection support for observation aggregation; RMC derives interaction confidence from this support to regulate track-query and memory updates.

\item Extensive experiments demonstrate improved tracking performance across challenging MOT benchmarks, together with more reliable association behavior, faster convergence, and improved data efficiency, while introducing only modest additional model complexity.

\end{itemize}

\section{Related Work}

The development of multi-object tracking (MOT) has been largely shaped by two representative paradigms: tracking-by-detection (TBD) and tracking-by-query (TBQ). This section reviews these two lines of work, with an emphasis on how interaction modeling has evolved from manually designed association to learnable query interaction, and why regularizing such interaction remains important for robust identity association.

\subsection{Tracking-by-Detection}

Tracking-by-detection has long dominated MOT due to its modular design, high efficiency, and interpretable tracking behavior. Classic methods such as SORT \cite{bewley2016simple} associate detections using IoU-based matching, establishing a simple yet effective pipeline based on Hungarian assignment. ByteTrack \cite{zhang2022bytetrack} further improves association robustness by exploiting low-confidence detections that may correspond to occluded objects, rather than discarding them before matching.

Following this line, subsequent studies have further explored how robust tracking can be maintained under ambiguous observations and association cues. OC-SORT \cite{cao2023observation} revisits motion modeling from an observation-centric perspective, reducing accumulated errors caused by unreliable prediction during lost periods. PD-SORT \cite{wang2025pd} incorporates pseudo-depth cues to alleviate occlusion-induced association ambiguity, while Hybrid-SORT \cite{yang2024hybrid} combines heterogeneous spatial cues, including confidence, height, and velocity direction, to improve association robustness. DCP-MOT \cite{hu2024dynamic} designs adaptive center-point sampling to avoid heavily occluded regions. Other studies~\cite{li2025advancing, zhang2026real} exploit relative spatial relationships to organize occlusion-aware association and recover lost trajectories. Detection-box area has also been used to organize cascaded matching~\cite{gu2024cascaded}.

Recent studies further explore more adaptive and progressive association mechanisms. TrackTrack \cite{shim2025focusing} refines association from the track perspective, while PlugTrack \cite{kim2026plugtrack} adaptively fuses classical Kalman filtering with data-driven motion predictors to handle both linear and nonlinear motion patterns. DiffMOT~\cite{lv2024diffmot} learns nonlinear motion prediction through a diffusion model, while GeneralTrack~\cite{qin2024towards} models visual and geometric point--region relations to improve generalization across tracking scenarios. AIPT~\cite{zhang2024aipt} further explores adaptive motion perception and information suppression to improve tracking under camera motion and ambiguous associations. OFTrack~\cite{song2025temporal} extends temporal modeling beyond individual frames through section-based object-flow estimation and temporal-coherence constraints, enabling association across multiple consecutive frames.
C-TWiX \cite{miah2025learning} learns coordinate-based affinity between tracklets. 
HGN~\cite{zhang2025heterogeneous} integrates heterogeneous association cues through semantic edges in a bipartite graph. PuTR~\cite{liu2025pure} represents trajectories as frame-ordered object sequences and uses structured attention masks to model short- and long-term associations in a separated online pipeline.
ConsistencyTrack \cite{jiang2025consistencytrack} reformulates joint detection and association as a progressive denoising process to improve tracking robustness.

Appearance maintenance and contextual representation learning provide complementary ways to improve association. CATrack~\cite{wang2025catrack} adapts appearance association and feature updates to target conditions, while HCgNet~\cite{li2024hcgnet} combines hierarchical context aggregation with tracklet-guided object refinement. LDMOT~\cite{zhong2026robust} further combines local feature matching with dynamic feature exchange to improve discrimination between similar targets.

Overall, TBD methods demonstrate the practical value of explicit association structures and physical cues. However, in many pipelines, these structures are encoded in matching costs or specialized association procedures, while feature aggregation and temporal state updating are handled through separate mechanisms. This separation leaves room for a unified interaction formulation in which task-specific regularities jointly guide observation integration and identity-state evolution.

\subsection{Tracking-by-Query}

With the development of Transformer architectures, tracking-by-query has emerged as an end-to-end alternative to heuristic association pipelines. MOTR \cite{zeng2022motr} introduces continuous track queries to represent tracked instances and update them over time. MO3TR \cite{zhu2022looking} models object interactions through spatial and temporal Transformers. 
% TrackFormer

While TBQ methods provide a unified formulation for localization, association, and temporal updating, the query interaction process introduces new optimization and association challenges. MeMOT \cite{cai2022memot} and MeMOTR \cite{gao2023memotr} introduce memory mechanisms to preserve long-term target information and strengthen temporal association. MOTIP \cite{gao2025multiple} decouples the framework and reformulates association as in-context identity prediction. CO-MOT \cite{yan2025comot} introduces a coopetition label assignment strategy to enhance the optimization of detection queries and alleviate the imbalance between detection and tracking. MOTRv2 \cite{zhang2023motrv2} bootstraps end-to-end tracking with detector-generated proposals to ease the task conflict. TBDQ-Net \cite{jia2026tracking} integrates a frozen detector with a lightweight query associator, forming an efficient intermediate paradigm between TBD and TBQ.

Recent studies further examine the internal representation and interaction behavior of query-based tracking. FDTA \cite{shao2025detection} observes that DETR-style object embeddings may exhibit high inter-object similarity during association, which can induce ambiguous interaction responses. LA-MOTR \cite{wang2025motr} employs spatially guided learnable attention to predict explicit object–tracklet associations. TCEI \cite{guo2026dual} improves online robustness with a human-inspired test-time calibration framework. In addition to standard MOT, open-world tracking studies such as GIT \cite{hu2022global} indicate that robust long-term tracking often requires interaction mechanisms beyond short-term affinity matching.

Taken together, these works show that modern MOT has moved beyond isolated pairwise matching toward detection-track cooperation, memory modeling, and prior-guided adaptation. Building on these advances, PI-MOT incorporates complementary structural regularities into learnable query interaction, jointly regulating spatial affinity, association competition, and reliability-aware track-state updating without introducing separate heuristic matching procedures.

\begin{figure*}[pos=t]
    \centering
    \includegraphics[width=1.0\textwidth]{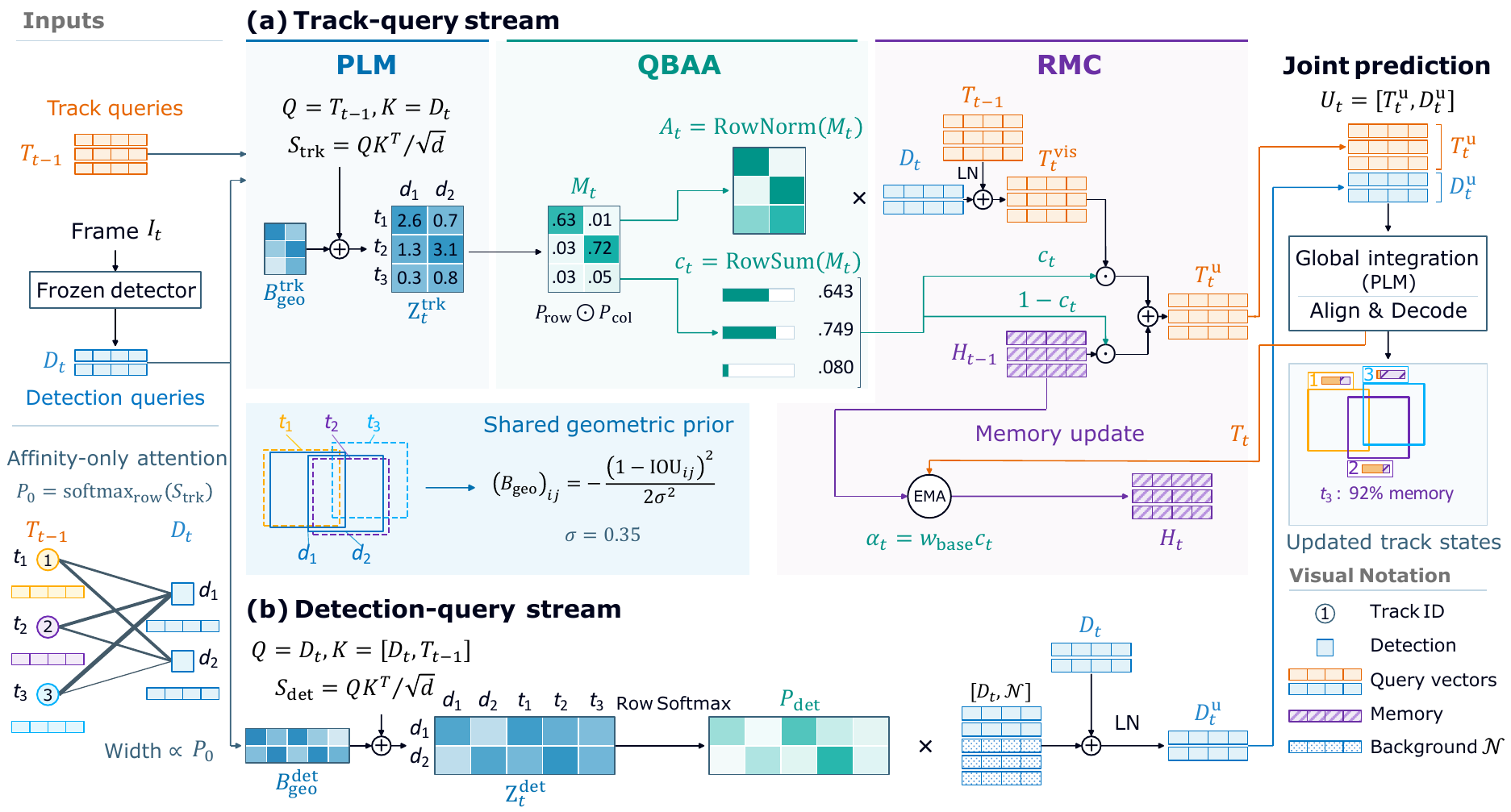}
    \caption{
    \textbf{Overall architecture of PI-MOT.} (a) PLM promotes spatial consistency in learned query affinity, and QBAA alleviates ambiguous competition through bidirectional reweighting. The resulting joint interaction matrix $\mathbf{M}_t$ provides aggregation weights $\mathbf{A}_t$ for observation aggregation and interaction confidence $\mathbf{c}_t$ for RMC to update track queries and historical memory. (b) Detection queries use spatially modulated attention. The updated track and detection queries are then integrated for final prediction.
    }
    \label{fig:overall}
    \vspace{-10pt}
\end{figure*}

\section{Methodology}
\label{sec:method}

\subsection{Overall Architecture}
\label{sec:overview}

PI-MOT is an end-to-end multi-object tracking framework that models identity association as a learnable information routing process guided by fundamental tracking regularities. PLM promotes spatial consistency in learned affinity, which QBAA uses to evaluate reciprocal support and alleviate ambiguous competition. RMC uses the resulting interaction confidence to balance current observations and historical memory in track-query updates and to adjust the memory update rate.

At frame $t$, $\mathbf D_t\in\mathbb{R}^{M\times d}$ denotes $M$ current detection queries, while $\mathbf T_{t-1}, \mathbf H_{t-1}\in\mathbb{R}^{N\times d}$ represent $N$ track queries and their historical memory propagated from the last frame; $d$ is the feature dimension. Detection features are sampled from the frozen detector’s feature maps, and box coordinates are sinusoidally encoded for attention. As shown in Fig.~\ref{fig:overall}(a), the track stream aggregates $\mathbf D_t$ into $\mathbf T_{t-1}$ and incorporates $\mathbf H_{t-1}$ through RMC. The detection stream in Fig.~\ref{fig:overall}(b) supports new-target discovery and duplicate suppression. The updated queries are concatenated as $\mathbf U_t=[\mathbf T_t^u, \mathbf D_t^u]$ for global integration with PLM.

To retain a favorable balance between tracking accuracy and computational efficiency, we use the decoupled
end-to-end framework TBDQ-Net \cite{jia2026tracking} for the surrounding pipeline, including encoding detector outputs as
queries and final content–position alignment and decoding.

%\mathbf{H}

\subsection{Prior-Informed Interaction Regularization}
\label{sec:interaction}

We detail the track-query computation in Fig.~\ref{fig:overall}(a), starting with the construction of track–detection interaction logits $\mathbf Z^{\mathrm{trk}}_t$. We then derive aggregation weights $\mathbf A_t$ for integrating current detections and interaction confidence $\mathbf c_t$ for regulating track-query and memory updates.

\subsubsection{Proximity-Guided Logit Modulation (PLM)}

PLM encourages spatially consistent observation aggregation with overlap-based modulation of learned query
affinity. For query $i$ and key $j$, let $\operatorname{IoU}_{ij}$ denote the intersection-over-union between their associated bounding boxes, obtained from the propagated track predictions and current detector outputs, respectively. We define
\begin{equation}
  d_{ij}=1-\operatorname{IoU}_{ij},\qquad
  (\mathbf B_{\mathrm{geo}})_{ij}=-\frac{d_{ij}^{2}}{2\sigma^{2}},
  \label{eq:geo}
\end{equation}
where $\sigma$ controls the strength of the geometric modulation. The interaction logits used to update track queries are:
\begin{equation}
  \mathbf Z_t^{\mathrm{trk}}=
  \frac{\mathbf T_{t-1}\mathbf D_t^{\top}}{\sqrt d}
  +\mathbf B_{\mathrm{geo}}^{\mathrm{trk}}.
  \label{eq:track_logit}
\end{equation}
The geometric term downweights low-overlap pairs while favoring pairs with greater overlap during observation aggregation. Its additive form preserves the contribution of learned query affinity, so that the interaction weights reflect both high-dimensional feature relationships and explicit spatial consistency.

\subsubsection{Quasi-Bijective Attentional Alignment (QBAA)}

When several track queries compete for the same detection, unconstrained observation aggregation can inject similar features into multiple identity representations, weakening their distinctiveness and increasing the risk of identity switches. QBAA alleviates this ambiguity through soft, differentiable reweighting of track–detection pairs according to their reciprocal support, without enforcing an exact one-to-one assignment.

Starting from the spatially modulated logits $\mathbf Z_t^{\mathrm{trk}}$, QBAA constructs the joint interaction matrix $\mathbf M_t$ by combining row- and column-wise softmax weights:
\begin{equation}
\begin{aligned}
  \mathbf P_{\mathrm{row}}&=\operatorname{softmax}_{\mathrm{row}}(\mathbf Z_t^{\mathrm{trk}}),\\
  \mathbf P_{\mathrm{col}}&=\operatorname{softmax}_{\mathrm{col}}(\mathbf Z_t^{\mathrm{trk}}),\\
  \mathbf M_t&=\mathbf P_{\mathrm{row}}\odot\mathbf P_{\mathrm{col}},
\end{aligned}
\label{eq:qbaa}
\end{equation}
where $\odot$ denotes element-wise multiplication. $\mathbf P_{\mathrm{row}}$ expresses each track’s preference over detections, and $\mathbf P_{\mathrm{col}}$ each detection’s preference over tracks. Their product measures reciprocal support within each track–detection pair. For example, track $i$ may assign a large row-wise weight to detection $j$, but a stronger response from another track reduces $(\mathbf{P}_{\mathrm{col}})_{ij}$ and hence the joint weight $(\mathbf{M}_{\mathrm{t}})_{ij}$.

For observation aggregation, we normalize each row of the joint interaction matrix $\mathbf{M}_t$ to obtain aggregation
weights $\mathbf{A}_t$:
\begin{equation}
  (\mathbf A_t)_{ij}=\frac{(\mathbf M_t)_{ij}}{\sum_k(\mathbf M_t)_{ik}},\qquad
  \mathbf T_t^{\mathrm{vis}}=\operatorname{LN}\!\left(\mathbf T_{t-1}+\mathbf A_t\mathbf D_t\right).
  \label{eq:aggregation}
\end{equation}
Here, $\mathbf A_t$ weights the current detection features, and $\mathbf T_t^{\mathrm{vis}}$ is the track representation after observation aggregation. The operator LN denotes LayerNorm. For brevity, the equations and Fig.~\ref{fig:overall} omit attention projections, positional
additions, dropout, and the residual feed-forward sublayers with their second LayerNorm.

\subsubsection{Reliability-Guided Memory Compensation (RMC)}
\label{sec:rmc}

RMC uses interaction confidence to regulate both track-query and memory updates, increasing reliance on historical information when current reciprocal support is weak. 
To assess the overall support strength, we define interaction confidence $\mathbf c_t$ as the row sum of $\mathbf M_t$:
\begin{equation}
  (\mathbf c_t)_i=\sum_j(\mathbf M_t)_{ij}
  =\sum_j(\mathbf P_{\mathrm{row}})_{ij}(\mathbf P_{\mathrm{col}})_{ij}.
  \label{eq:reliability}
\end{equation}

A track receives high-confidence interaction when its preferred detections also strongly favor it over competing tracks. Competition for those detections weakens this reciprocal support and lowers the confidence. We first use $\mathbf c_t$, broadcast across feature dimensions, to balance the visually updated track representation $\mathbf T_t^{\mathrm{vis}}$ with historical memory $\mathbf H_{t-1}$:
\begin{equation}
  \mathbf T_t^{u}=\mathbf c_t\odot\mathbf T_t^{\mathrm{vis}}
  +(1-\mathbf c_t)\odot\mathbf H_{t-1}.
  \label{eq:track_update}
\end{equation}
Higher confidence increases the contribution of $\mathbf T_t^{\mathrm{vis}}$, while lower confidence favors historical memory $\mathbf H_{t-1}$.

After global interaction and decoding, the final track representations $\mathbf T_t$ are used to update historical memory. We use the interaction confidence $\mathbf c_t$ to modulate the update rate of an exponential moving average:
\begin{equation}
  \mathbf H_t=(w_{\mathrm{base}}\mathbf c_t)\odot\mathbf T_t
  +(1-w_{\mathrm{base}}\mathbf c_t)\odot\mathbf H_{t-1},
  \label{eq:memory_update}
\end{equation}
where $w_{\mathrm{base}}$ is a fixed base update rate that stabilizes temporal accumulation.

\subsection{Detection Query Update and Prediction}

Detection queries discover new objects while learning to suppress responses that duplicate existing tracks. In this stream, $\mathbf D_t$ serves as the queries and $[\mathbf D_t,\mathbf T_{t-1}]$ as the keys. PLM supplies the bias computed from the boxes
associated with these query–key pairs:
\begin{equation}
  \mathbf Z_t^{\mathrm{det}}=
  \frac{\mathbf D_t[\mathbf D_t,\mathbf T_{t-1}]^{\top}}{\sqrt d}
  +\mathbf B_{\mathrm{geo}}^{\mathrm{det}}.
  \label{eq:det_logit}
\end{equation}
The values are $[\mathbf D_t,\mathcal N]$, where $\mathcal N\in\mathbb{R}^{N\times d}$ contains auxiliary background values. These occupy the positions of the $N$ track keys, absorbing redundant responses associated with existing tracks. With $\mathbf P_{\mathrm{det}}=\operatorname{softmax}_{\mathrm{row}}(\mathbf Z_t^{\mathrm{det}})$, as shown in Fig.~\ref{fig:overall}(b), the update is
\begin{equation}
  \mathbf D_t^u=\operatorname{LN}\!\left(
  \mathbf D_t+\mathbf P_{det}
  [\mathbf D_t,\mathcal N]\right).
  \label{eq:det_update}
\end{equation}
We concatenate the updated track and detection queries as $\mathbf U_t=[\mathbf T_t^u,\mathbf D_t^u]$ and apply a spatially modulated global interaction to exchange contextual information between the two streams:
\begin{equation}
\begin{aligned}
  \mathbf Z_t^{u}&=\frac{\mathbf U_t\mathbf U_t^{\top}}{\sqrt d}
  +\mathbf B_{\mathrm{geo}}^{u},\\
  \mathbf U_t^{\mathrm{out}}&=\operatorname{LN}\!\left(
  \mathbf U_t+\operatorname{softmax}_{\mathrm{row}}(\mathbf Z_t^{u})\mathbf U_t
  \right).
\end{aligned}
\label{eq:global}
\end{equation}
Here, $\mathbf B_{\mathrm{geo}}^{u}$ uses the boxes associated with the concatenated queries. Content–position alignment (CPA) then uses global image features to reconcile the updated query features with their positional states, followed by decoding into boxes and classification scores. These supporting operations \cite{jia2026tracking} underpin the effectiveness of the decoupled end-to-end tracking framework and form the “Align \& decode” block in Fig.~\ref{fig:overall}. The final track states are propagated as query inputs for the next frame.

\subsection{Training and Online Inference}
\label{sec:prediction}

Training uses dual-label assignment \cite{jia2026tracking}. Detection-query predictions are assigned to newly appearing objects by bipartite matching, while track queries receive temporal identity supervision through inherited assignments.
The training objective combines focal classification loss, L1 box regression loss, and generalized IoU loss:
\begin{equation}
  \mathcal L=\lambda_{\mathrm{cls}}\mathcal L_{\mathrm{cls}}
  +\lambda_{L1}\mathcal L_{\mathrm{L1}}
  +\lambda_{\mathrm{giou}}\mathcal L_{\mathrm{giou}},
  \label{eq:loss}
\end{equation}
with $(\lambda_{\mathrm{cls}},\lambda_{\mathrm{L1}},\lambda_{\mathrm{giou}})=(2,5,2)$. 

We introduce Endogenous Track Augmentation (ETA) to construct hard negatives from the model’s own track predictions. During training, an active-track prediction whose box has IoU below 0.5 with its assigned ground-truth box is relabeled as background, and its corresponding track query is propagated to the next frame as a hard negative. These hard negatives encourage the model to suppress inaccurately localized track predictions.

During online inference, detection-query predictions with classification scores above 0.5 initialize new tracks, while active track queries above the same threshold are propagated. Tracks falling below the threshold are marked inactive and removed after 20 consecutive inactive frames.

\begin{algorithm}[t]
\caption{Training and Inference Pipeline of PI-MOT}
\label{alg:pimot}
\begin{algorithmic}[1]
\Require Video $\{I_t\}_{t=1}^{L}$; frozen detector
$\mathrm{Det}$; $\tau_q=0.5$; $w_{\mathrm{base}}=0.3$; $\tau_n=0.5$; 
max inactive age $L_{\max}=20$
\Ensure Tracklets $\mathcal T$

\State Initialize $\mathcal T\gets\emptyset$,
$\mathcal H\gets\emptyset$,
$\sigma_0^{\mathrm{tr}}\gets\emptyset$

\For{$t=1,\ldots,L$}
    \State $(F_t,P_t,D_t,B_t^{\mathrm{det}},s_t)
    \gets\Call{DetectAndEncode}{I_t}$
    \Statex \hspace{\algorithmicindent}%
    \parbox[t]{0.85\linewidth}{\footnotesize
    $F_t$: image features; $P_t$: positional encodings;
    $D_t$: detection queries; $B_t^{\mathrm{det}}$: detection boxes;
    $s_t$: scores}
    \State Select qualified detection queries using $\tau_q$
    \State Retrieve $T_{t-1},B_{t-1}^{\mathrm{trk}},H_{t-1}$
    from $\mathcal T,\mathcal H$
    \Statex \hspace{\algorithmicindent}%
    \parbox[t]{0.85\linewidth}{\footnotesize
    $B_{t-1}^{\mathrm{trk}}$:
    previously predicted track boxes}
    
    \If{$|T_{t-1}|>0$ \textbf{and} qualified detection queries exist}
        \State Prepare auxiliary background candidates $\mathcal N_t$
        \State $Z_t^{\mathrm{trk}}\gets
        \Call{PLM}{T_{t-1},D_t,B_{t-1}^{\mathrm{trk}},B_t^{\mathrm{det}}}$
        \State $(M_t,A_t,c_t)\gets
        \Call{QBAA}{Z_t^{\mathrm{trk}}}$
        \State $T_t^{\mathrm{vis}}\gets
        \Call{Aggregate}{T_{t-1},A_t,D_t}$
        \State $T_t^u\gets
        \Call{RMC}{T_t^{\mathrm{vis}},H_{t-1},c_t}$
        \State $D_t^u\gets
        \Call{DetectionStream}{D_t,T_{t-1},\mathcal N_t;\mathrm{PLM}}$
        \State $U_t\gets[T_t^u,D_t^u]$
        \State $U_t^{\mathrm{out}}\gets
        \Call{GlobalInteraction}{U_t;\mathrm{PLM}}$
        \State $(Q_t,Q_t^p)\gets
        \Call{CPA}{F_t,P_t,U_t^{\mathrm{out}}}$
    \Else
        \State $(Q_t,Q_t^p)\gets
        ([D_t,T_{t-1}],
        [B_t^{\mathrm{det}},B_{t-1}^{\mathrm{trk}}])$
    \EndIf

    \State $(\hat y_t,T_t)\gets
    \Call{PredictAndDecode}{F_t,P_t,Q_t,Q_t^p}$

    \If{prior-informed interaction was performed}
        \State $H_t\gets
        (w_{\mathrm{base}}c_t)\odot T_t
        +(1-w_{\mathrm{base}}c_t)\odot H_{t-1}$
    \Else
        \State $H_t\gets H_{t-1}$
    \EndIf

    \If{training}
        \State $\sigma_t^{\mathrm{det}}\gets
        \Call{Match}{\hat y_t^{\mathrm{det}},y_t^{\mathrm{new}}}$
        \State $\sigma_t^{\mathrm{tr}}\gets
        \sigma_{t-1}^{\mathrm{tr}}\cup\sigma_t^{\mathrm{det}}$
        \State Compute tracking loss and apply ETA
        \State Update supervised track queries
    \Else
        \State Update existing tracks and initialize new tracks
        using $\tau_n$
        \State Remove tracks inactive for more than $L_{\max}$ frames
    \EndIf
    \State Maintain historical memory for retained and new tracks
\EndFor
\State \Return $\mathcal T$
\end{algorithmic}
\end{algorithm}

\section{Experiments}

We evaluate PI-MOT on three challenging MOT benchmarks to assess overall tracking performance. Beyond that, we investigate data efficiency, training convergence, and computational overhead to assess the optimization benefits and practical cost of the introduced structural regularization. Ablation studies quantify the contributions of the proposed mechanisms, while mechanism analysis examines the behavior and applicability of structurally regularized interaction.

\subsection{Benchmarks and Metrics}
We evaluate PI-MOT on three challenging MOT benchmarks: DanceTrack \cite{sun2022dancetrack}, SportsMOT \cite{cui2023sportsmot}, and MOT20 \cite{dendorfer2020mot20}.
DanceTrack features targets with uniform appearances and complex non-linear motions, making it suitable for evaluating identity association under complex motion scenarios. SportsMOT further introduces rapid camera motion and similar player appearances, posing additional challenges for robust tracking. MOT20 focuses on extremely crowded pedestrian scenes with severe occlusions, providing a challenging benchmark for evaluating identity association in dense environments.
Unless otherwise specified, controlled analyses are conducted on the DanceTrack validation set, whereas comparisons with state-of-the-art methods are reported on the official test set of each benchmark. \textit{Throughout all experiments, we do not employ external data, such as CrowdHuman~\cite{shao2018crowdhuman}, for trajectory synthesis.} 

Following standard evaluation protocols, we report HOTA \cite{luiten2021hota}, DetA, AssA, IDF1 \cite{ristani2016performance}, and MOTA \cite{bernardin2008evaluating} on DanceTrack and SportsMOT. For MOT20, we report HOTA, MOTA, IDF1, FP, FN, and IDSW following the MOTChallenge protocol.

\subsection{Implementation Details}
\label{sec:imp_details}

Our model is implemented in the PyTorch framework and trained on 2 NVIDIA RTX 4090 GPUs. The surrounding
tracking pipeline and common training settings follow TBDQ-Net \cite{jia2026tracking} for its favorable balance between
tracking accuracy and computational efficiency. Specifically, we utilize the publicly released YOLOX \cite{ge2021yolox} model, pretrained by ByteTrack \cite{zhang2022bytetrack}, as the frozen detector. The image processing and sequence sampling strategies follow standard practices commonly adopted in recent TBQ trackers such as MOTRv2 \cite{zhang2023motrv2}, CO-MOT \cite{yan2025comot}, and TBDQ-Net \cite{jia2026tracking}, including random scaling, random cropping, and contiguous video clip sampling for temporal continuity.

The initial learning rate is set to $1.2\times10^{-4}$ for all datasets. On DanceTrack, the model is trained for $12$ epochs, with the learning rate decayed by a factor of $10$ at epochs $6$ and $10$. To properly accommodate the varying data volumes, the training schedules are correspondingly extended for SportsMOT ($20$ epochs total, decayed at $12$ and $16$) and MOT20 ($30$ epochs total, decayed at $20$ and $25$). Core hyperparameters, including the base modulation weight $w_{base}=0.3$ (Sec.~\ref{sec:rmc}), the 0.5 confidence threshold, and the 20-frame inactive patience (Sec.~\ref{sec:prediction}), are shared across all evaluation datasets without scene-specific tuning.

Due to the substantially smaller training scale and limited scene diversity of MOT20 (merely $8.9$K frames across $4$ sequences, representing nearly $5\times$ less volumes and $10\times$ fewer scenes than DanceTrack), training TBQ trackers poses a high risk of overfitting \cite{zeng2022motr,yan2025comot,jia2026tracking}.
To mitigate data scarcity, we apply temporal augmentations exclusively for MOT20: expanding the training clip length and introducing a random dropout of track instances with a probability of $0.1$. These MOT20-specific augmentation settings are inherited from TBDQ-Net, with no further dataset-specific tuning.

Controlled Baseline. For the controlled analyses, we construct an affinity-driven baseline within the same tracking pipeline as PI-MOT, without PLM, QBAA, or RMC. The frozen detector, query encoding, prediction pipeline, and corresponding training settings are shared between the two models. We use this baseline in the learning-efficiency and component analyses below. Its specific interaction and state-update settings are detailed in Sec.~\ref{sec:exps_abla}.

\subsection{Comparison with State-of-the-Art Methods}
\label{sec:cmp_sota}

\begin{table*}[t]
\centering
\footnotesize
\caption{\textbf{Comparison on the
DanceTrack test set.} TBD methods are listed above TBQ methods. Dash symbols (--) indicate the data is unavailable. $\uparrow$ denotes that higher values are preferable. Methods marked with an asterisk ($^{*}$) indicate the use of the identical pretrained YOLOX detector. \textbf{Bold} and \uwave{underwaved} values represent the best and second-best performance in each category, respectively. Subscripts (e.g., $+2.5$) report the performance gains of PI-MOT over the competitive methods for certain metrics.}
\label{exps:cmp_dance}
\setlength{\tabcolsep}{3.6pt}
\begin{tabular}{c|c|ccccc}
\toprule
Methods & Year  & \textbf{HOTA}$\uparrow$ & DetA$\uparrow$ & AssA$\uparrow$ & MOTA$\uparrow$ & IDF1$\uparrow$ \\ \hline
ByteTrack$^{*}$   \cite{zhang2022bytetrack}           & 2022                 & 47.7 & 71.0 & 32.1 & 89.6 & 53.9 \\
OC-SORT$^{*}$    \cite{cao2023observation}            & 2023  & 55.1 & 80.3 & 38.3 & 92.0 & 54.6 \\

ETTrack$^{*}$     \cite{han2025ettrack}  & 2024        & 56.4 & 81.7  & 39.1 & 92.2  & 57.5 \\
Hybrid-SORT$^{*}$      \cite{yang2024hybrid}      &  2024                          & 65.7 & --    & --    & 91.8 & 67.4 \\ %\hline

MambaMOT$^{*}$    \cite{Huang2024MambaMOTSM}            &  2025             & 56.1 & 80.8  & 39.0  & 90.3 & 54.9 \\
C-TWiX$^{*}$      \cite{miah2025learning}    &  2025                        & 62.1 & 81.8 & 47.2 & 91.4 & 63.6 \\  

REAL--SORT$^{*}$~\cite{zhang2026real} & 2026 & 64.0 & 81.9 & 47.8 & 92.3 & 65.2 \\

LDMOT$^{*}$~\cite{zhong2026robust} & 2026 & 65.8 &  81.6 & 53.3 & 91.4 & 68.4 \\ \midrule 

MOTR                       \cite{zeng2022motr}        &  2022               & 54.2 & 73.5 & 40.2 & 79.7 & 51.5 \\
MeMOTR                     \cite{gao2023memotr}       &  2023                & 68.5 & 80.5 & 58.4 & 89.9 & 71.2 \\ 

MOTRv2$^{*}$                     \cite{zhang2023motrv2}     & 2023               & 69.9 & 83.0 & 59.0 & 91.9 & 71.7 \\ 

SambaMOTR \cite{segu2025samba} & 2025 & 67.2 & 78.8 & 57.5 & 88.1 & 70.5 \\

CO-MOT  \cite{yan2025comot}  & 2025  & 69.4 & 82.1 & 58.9 & 91.2 & 71.9 \\ 

MOTIP                      \cite{gao2025multiple}     & 2025              & 70.0 & 80.8$_{+2.5}$ & \textbf{60.8} & 91.0$_{+1.4}$ & \textbf{75.1} \\ 

LA--MOTR \cite{wang2025motr} & 2025 & \textbf{71.1} & \textbf{84.4} & 59.0$_{+1.1}$ & \textbf{93.5} & 71.8$_{+2.2}$ \\

DecoderTracker \cite{liao2026decodertracker} & 2026 & 54.5 & 79.0 & 35.6 & 88.1 & 54.5 \\
TBDQ--Net$^{*}$ \cite{jia2026tracking}       & 2026  & 69.4 & 80.8$_{+2.5}$ & 59.8 & 89.5$_{+2.9}$ & 73.4 \\ \midrule

\rowcolor{ourscolor} $\mathrm{\textbf{PI-MOT}}^{*}$  & ours & \uwave{70.7}  &  \uwave{83.3}  &  \uwave{60.1} & \uwave{92.4}   & \uwave{74.0} \\
\bottomrule
\end{tabular}
\vspace{-4pt}
\end{table*}

\begin{table*}[pos=t]
\centering
\footnotesize
\setlength{\tabcolsep}{4pt} % 适当放宽列间距，因为总列数减少了
\caption{\textbf{Comparison on the SportsMOT test set.} The evaluation is stratified into two regimes: Setting I utilizes only the training set (\textit{train}), while Setting II combines the training and validation sets (\textit{train+val}). All other notations are consistent with those presented in Tab. \ref{exps:cmp_dance}.} 
\label{exps:cmp_sports}
\begin{tabular}{l|c|ccccc}
\toprule
Methods & Year & \textbf{HOTA}$\uparrow$ & DetA$\uparrow$ & AssA$\uparrow$ & MOTA$\uparrow$ & IDF1$\uparrow$ \\ 
\midrule
\multicolumn{7}{c}{\textcolor{gray}{\textit{Setting I: train}}} \\
ByteTrack$^{*}$ \cite{zhang2022bytetrack}  & 2022 & 62.8 & 77.1 & 51.2 & 94.1 & 69.8 \\
OC-SORT$^{*}$  \cite{cao2023observation}   & 2023 & 71.9 & 86.4 & 59.8 & 94.5 & 72.2 \\
ETTrack$^{*}$    \cite{han2025ettrack}  & 2024 & 72.2 & \uwave{86.9} & 60.1 & 94.9 & 72.5 \\
MambaMOT$^{*}$    \cite{Huang2024MambaMOTSM} & 2025 & 71.3 & 86.7 & 58.6 & 94.9 & 71.1 \\ 
\midrule
MeMOTR \cite{gao2023memotr} & 2023 & 70.0 & 83.1 & 59.1 & 91.5 & 71.4 \\
SambaMOTR \cite{segu2025samba} & 2025 & 69.8 & 82.2 & 59.4 & 90.3 & 71.9 \\
MOTIP \cite{gao2025multiple} & 2025 & 71.9 & 83.4$_{+4.0}$ & 62.0 & 92.9$_{+2.4}$ & \uwave{75.0} \\ 
LA-MOTR \cite{wang2025motr} & 2025 & \uwave{72.4} & 86.6 & 61.8$_{+2.7}$ & \textbf{95.6} & 73.3$_{+2.5}$ \\ 
TBDQ-Net$^{*}$ \cite{jia2026tracking} & 2026 & 72.1 & 82.8$_{+4.6}$ & \uwave{62.8} & 89.5$_{+5.8}$ & 72.9 \\ 
\rowcolor{ourscolor} \textbf{PI-MOT}$^{*}$ & ours & \textbf{75.0} & \textbf{87.4} & \textbf{64.5} & \uwave{95.3} & \textbf{75.8} \\ 
\midrule
\multicolumn{7}{c}{\textcolor{gray}{\textit{Setting II: train+val}}} \\
ByteTrack$^{*}$ \cite{zhang2022bytetrack}  & 2022 & 64.1 & 78.5 & 52.3 & 95.9 & 71.4 \\
OC-SORT$^{*}$  \cite{cao2023observation}   & 2023 & 73.7 & 88.5 & 61.5 & 96.5 & 74.0 \\
MixSORT$^{*}$  \cite{cui2023sportsmot}     & 2023 & 74.1 & 88.5 & 62.0 & 96.5 & 74.4 \\
ETTrack$^{*}$    \cite{han2025ettrack}  & 2024 & 74.3 & \uwave{88.8} & 62.1 & \uwave{96.8} & 74.5 \\ \midrule
MOTIP \cite{gao2025multiple} & 2025 & 75.2 & 86.5$_{+2.7}$ & 65.4 & 96.1$_{+0.9}$ & \textbf{78.2} \\ 
TBDQ-Net$^{*}$ \cite{jia2026tracking} & 2026 & \uwave{75.5} & 85.9$_{+3.3}$ & \uwave{66.5} & 92.9$_{+4.1}$ & 76.1 \\ 
\rowcolor{ourscolor} \textbf{PI-MOT}$^{*}$ & ours & \textbf{77.0} & \textbf{89.2} & \textbf{66.6} & \textbf{97.0} & \uwave{77.6} \\ 
\bottomrule
\end{tabular}
\vspace{-4pt}
\end{table*}

\paragraph{Comparison on DanceTrack} 
As presented in Tab.~\ref{exps:cmp_dance}, PI-MOT achieves top-tier performance across all evaluation metrics on the DanceTrack benchmark. Compared with recent TBQ trackers, our method attains a favorable balance between detection accuracy and identity association, achieving 83.3 DetA and 60.1 AssA simultaneously. In particular, compared with MOTIP, which excels in identity association, PI-MOT maintains competitive association performance while improving DetA/MOTA by +2.5/+1.4. Conversely, compared with LA-MOTR, which exhibits strong detection performance, PI-MOT achieves comparable detection accuracy while improving AssA/IDF1 by +1.1/+2.2. These results indicate that PI-MOT achieves a favorable balance between detection accuracy and identity association under highly ambiguous motion scenarios.

\paragraph{Comparison on SportsMOT}
Tab.~\ref{exps:cmp_sports} presents the evaluation on the SportsMOT benchmark, which involves rapid camera motion and highly similar player appearances. Under these challenging scenarios, PI-MOT achieves consistently strong tracking performance across both training settings. In the \textit{train} setting, PI-MOT achieves the best overall performance with 75.0 HOTA, 87.4 DetA, and 64.5 AssA, surpassing previous TBQ trackers by clear margins. In the \textit{train+val} setting, PI-MOT further improves to 77.0 HOTA and achieves the best performance across nearly all evaluation metrics. 

Notably, in the \textit{train} setting, PI-MOT already rivals or exceeds several methods trained with the larger \textit{train+val} split. This observation is consistent with its favorable learning efficiency, which is further examined through controlled experiments in Sec.~\ref{sec:data_efficiency}.

\begin{table*}[pos=!t]
\centering
\footnotesize
\caption{\textbf{Comparison on the MOT20 test set.} Scores marked in \textbf{bold} and \uwave{underwave} are the best and second-best among TBQ methods. All other notations are consistent with those presented in Tab.~\ref{exps:cmp_dance}.} 
\label{exps:cmp_mot20}
\begin{tabular}{c|c|cccccc}
\toprule
Methods                     &  Year      & \textbf{HOTA}$\uparrow$ & MOTA$\uparrow$ & IDF1$\uparrow$ & FP$\downarrow$ & FN$\downarrow$ & IDSW$\downarrow$ \\ \hline
ByteTrack$^{*}$ \cite{zhang2022bytetrack} & 2022 & 61.3 & 77.8 & 75.2 & 26249 &	87594 & 1223 \\
OC-SORT$^{*}$ \cite{cao2023observation} & 2023 & 62.4 & 75.9 & 76.4 & 20,218 & 103,791 & 938 \\
MSPNet \cite{zheng2024motion}                    & 2024       & --      & 75.9 & 74.4  & 26,532  & 96,909  & 1,194  \\
GLMB   \cite{van2024visual}                        & 2024       & 54.2    & 67.7 & 67.3 & 29,597  & 134,534 &  2,911  \\

OFTrack$^{*}$ \cite{song2025temporal} & 2025 & 61.9 & 75.3 & 74.7 & -- & -- & -- \\

OTTrack$^{*}$ \cite{li2025advancing} & 2025 & 62.3 & 76.1 & 76.2 & 20,700 & 101,900 & 1070 \\

PlugTrack$^{*}$ \cite{kim2026plugtrack} & 2026 & 61.8 & 76.4 & 75.2 & -- & -- & -- \\
ASDTracker$^{*}$ \cite{wang2026asdtracker} & 2026 & 62.5 & 76.5 & 76.3 & -- & 88,125 & 1,247 \\ \midrule \midrule

MeMOT               \cite{cai2022memot}          & 2022     & 54.1 & 63.7 & 66.1  & 47,882 & 137,983 & 1,938 \\
CO-MOT \cite{yan2025comot} & 2025 & 57.5 & 60.1 & 70.5 & -- & -- & -- \\  
TBDQ-Net$^{*}$ \cite{jia2026tracking} &  2026  & \uwave{60.2} & \uwave{72.2} & \uwave{72.9} & \uwave{30,950} & \uwave{111,869} & \uwave{1,200} \\ 

\rowcolor{ourscolor} $\mathrm{\textbf{PI-MOT}}^{*}$ & ours & \textbf{62.5} & \textbf{73.8} & \textbf{76.6} & \textbf{25,717} & \textbf{108,792} & \textbf{1,147} \\  \bottomrule
\end{tabular}
\vspace{-6pt}
\end{table*}

\begin{table*}[pos=!ht]
\caption{\textbf{Data efficiency under reduced training data.}
Validation results compare PI-MOT with the affinity-driven baseline on the DanceTrack validation set; test results report PI-MOT on the official test set. Subscripts indicate gains over the baseline at the same data
proportion.}
\label{exps:abla_data}
\centering
\footnotesize
\setlength{\tabcolsep}{3.8pt}
\begin{tabular}{c||ccc|ccc||ccc}
\toprule
\multirow{2}{*}{Data (\%)}
& \multicolumn{3}{c|}{Baseline (val)}
& \multicolumn{3}{c||}{PI-MOT (val)}
& \multicolumn{3}{c}{PI-MOT (test)} \\
\cmidrule(lr){2-4} \cmidrule(lr){5-7} \cmidrule(lr){8-10}
& HOTA & DetA & AssA
& HOTA & DetA & AssA
& HOTA & DetA & AssA \\
\midrule
100 & 59.3 & 76.2 & 46.4 & $66.9_{+7.6}$ & $79.2_{+3.0}$ & $56.7_{+10.3}$ & 70.7 & 83.3 &
60.1 \\
75  & 57.3 & 75.5 & 43.7 & $66.0_{+8.7}$ & $79.2_{+3.7}$ & $55.2_{+11.5}$ & 70.2 & 82.6 & 59.7 \\
50  & 55.5 & 73.1 & 42.3 & $65.2_{+9.7}$ & $78.2_{+5.1}$ & $54.6_{+12.3}$ & 68.7 & 82.1 & 57.6 \\
25  & 50.3 & 67.3 & 37.8 & $64.0_{+13.7}$ & $77.7_{+10.4}$ & $52.9_{+15.1}$ & 67.6 & 81.9 & 55.8 \\
\bottomrule
\end{tabular}
\vspace{-2pt}
\end{table*}

\paragraph{Comparison on MOT20}
Tab.~\ref{exps:cmp_mot20} reports the evaluation on the extremely crowded MOT20 benchmark, where comparatively few TBQ trackers report competitive results due to the dense pedestrians and limited scene diversity. PI-MOT achieves the best overall performance among TBQ methods across all reported metrics. In particular, compared with TBDQ-Net, PI-MOT improves HOTA from 60.2 to 62.5 and IDF1 from 72.9 to 76.6, while simultaneously reducing FP, FN, and IDSW. These results demonstrate that the proposed structural regularization generalizes effectively from highly dynamic motion scenarios to densely crowded environments with limited data.

\subsection{Learning and Computational Efficiency}
\label{sec:data_efficiency}

Beyond final tracking accuracy, we examine data efficiency and convergence speed, together with the computational cost of the complete framework.

\paragraph{Analysis of Data Efficiency} 
To further evaluate the data efficiency of PI-MOT, we compare the affinity-driven baseline and the full framework under varying training data proportions on DanceTrack. 
For each proportion,
we retain the first 25\%, 50\%, 75\%, or 100\% of frames from every training sequence, preserving sequence
coverage and temporal order. Both configurations use the same subset and the same total number of training
epochs. Each epoch traverses the corresponding subset once, so the actual number of updates per epoch decreases
with the subset size.

As summarized in Tab.~\ref{exps:abla_data}, PI-MOT maintains competitive tracking performance even under substantially reduced supervision. Notably, when trained with only 25\% of the training data, PI-MOT still achieves 64.0 HOTA on the validation set and 67.6 HOTA on the official test set. Furthermore, the performance gap between the two models widens as the training data decreases, expanding from a 7.6-point lead in HOTA using the full dataset to 13.7 points under the 25\% setting, alongside a 15.1-point gain in AssA. These results suggest that introducing lightweight structural regularization reduces the reliance on extensive training data for learning complex motion patterns.

\begin{figure}[pos=t]
    \centering
    \footnotesize
    \includegraphics[width=0.45\textwidth]{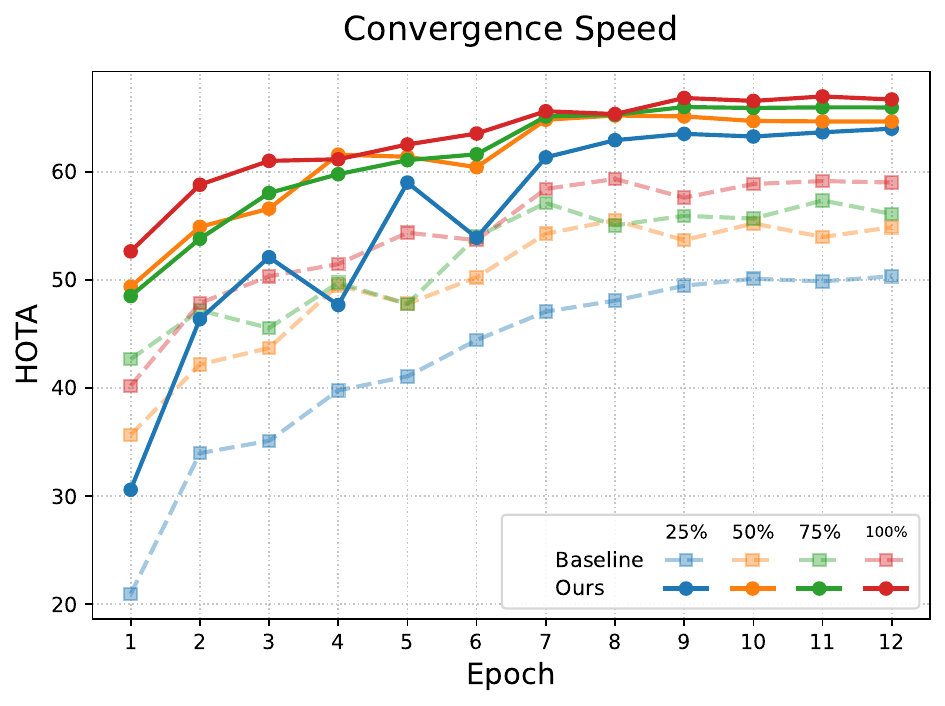}
    \caption{\textbf{Convergence speed comparison between the baseline and PI-MOT on the DanceTrack validation set.} Solid lines with circular markers represent PI-MOT, and dashed lines with square markers denote the baseline. Different colors indicate varying proportions of utilized training data.}
    \label{fig:vis_learning_speed}
\end{figure}

\paragraph{Analysis of Convergence Speed}
We further compare the convergence curves of PI-MOT and the affinity-driven baseline under varying training data proportions. Fig.~\ref{fig:vis_learning_speed} presents the validation HOTA across training epochs on DanceTrack.
The horizontal axis denotes epochs over the corresponding subset; comparisons across different proportions therefore also involve different numbers of optimization steps per epoch. At each data proportion, PI-MOT reaches competitive performance in fewer epochs than the baseline.
This advantage becomes more pronounced as the training data decreases. Notably, even with only 25\% of the training data, PI-MOT reaches the best HOTA level of the full-data baseline.
These observations suggest that structural regularization provides effective guidance during interaction learning, accelerating convergence even with limited supervision.

\begin{table}[t]
    \caption{\textbf{Computational efficiency comparison with representative TBQ methods.} `Params' count trainable weights in millions. Training time represents the duration on DanceTrack under the listed hardware configurations. Values are sourced from original publications, whereas the value marked by $\dagger$ is reproduced using the official implementation. FPS is evaluated on one NVIDIA RTX 4090 GPU with FP32 precision.}
    \label{exps:efficiency}
    \centering
    \footnotesize
    \setlength{\tabcolsep}{2.6pt}
    \begin{tabular}{c|c|cc|c}
    \toprule 
    Methods & Params (M)  & Training Time   & GPUs  & FPS   \\ \midrule
    MOTIP        \cite{gao2025multiple}   & 58.9    & 36h  & 8$\times$4090        & 14.4             \\
    CO-MOT \cite{yan2025comot}   & 41.7  & 125h$^{\dagger}$  & 2$\times$4090   & 14.9  \\ 
    LA-MOTR \cite{wang2025motr} & 50.3 & -- & -- & 12.9  \\ %\midrule
    TBDQ-Net \cite{jia2026tracking} & 8.5 & 22h & 2$\times$4090  & \textbf{22.0}   \\      
    \rowcolor{ourscolor} \textbf{PI-MOT (ours)} & 8.9  & 24.8h & 2$\times$4090 &  21.8   \\  \bottomrule
    \end{tabular}
    \vspace{-10pt}
\end{table}

\paragraph{Computational Efficiency Comparison} Tab.~\ref{exps:efficiency} summarizes the model size, training time, and inference speed of representative TBQ trackers. PI-MOT contains 8.9M trainable parameters, achieves an inference speed of 21.8 FPS, and requires 24.8 training hours on two RTX 4090 GPUs. Compared with larger TBQ frameworks such as MOTIP, CO-MOT, and LA-MOTR, PI-MOT remains substantially more compact while maintaining competitive inference efficiency. Together with the tracking results in Tabs.~\ref{exps:cmp_dance}-\ref{exps:cmp_mot20}, these results demonstrate a favorable balance between tracking performance and computational cost.

\subsection{Ablation Studies}
\label{sec:exps_abla}

\textit{Ablation Configurations.} We detail the component settings of the affinity-driven baseline introduced in
Sec.~\ref{sec:imp_details}. Relative to full PI-MOT, this configuration (1) removes the proximity-guided spatial bias $\mathbf{B}_{geo}$ from
all interaction matrices; (2) replaces QBAA’s bidirectional normalization with standard row-wise softmax attention;
and (3) replaces confidence-guided state compensation with uniform addition, $\mathbf{T}^u_t = \mathbf{T}^{vis}_t + \mathbf{H}_{t-1}$; historical memory is initialized with the query embedding when the track is confirmed and remains unchanged throughout its lifetime. This
configuration provides the starting point for the progressive component analysis in Tab.~\ref{exps:abla_each_part}.

\begin{table}[t]
\footnotesize
\caption{\textbf{Ablation study of the proposed components.}}
\label{exps:abla_each_part}
\centering
\begin{tabular}{ccc|ccc}
\toprule
PLM & QBAA & RMC &  HOTA & DetA & AssA \\ \hline
 &  &  & 59.3 & 76.2 & 46.4 \\
$\checkmark$ &  &  & 62.3 & 79.0 & 49.4 \\
$\checkmark$ & $\checkmark$ & & 65.4 & 78.9 & 54.3 \\
\rowcolor{ourscolor} $\checkmark$ & $\checkmark$ & $\checkmark$ & 66.9 & 79.2 & 56.7 \\ \bottomrule
\end{tabular}
\vspace{-4pt}
\end{table}

\paragraph{Component Analysis} 
To evaluate the contribution of each proposed module, we progressively integrate PLM, QBAA, and RMC into the baseline framework, as summarized in Tab.~\ref{exps:abla_each_part}. Introducing PLM immediately improves the overall tracking performance from 59.3 to 62.3 HOTA, with consistent gains in both detection and association accuracy. This result suggests that incorporating spatial regularization into affinity-driven interaction provides useful structural guidance in ambiguous motion scenarios. 
Building upon PLM, QBAA further improves HOTA from 62.3 to 65.4, with the gain mainly reflected in association accuracy (+4.9 AssA). This observation indicates that reducing ambiguous many-to-one competition is particularly beneficial for maintaining identity consistency during interaction.
Equipping the framework with RMC further improves the overall performance to 66.9 HOTA and 56.7 AssA. 
Since RMC primarily addresses interaction ambiguity induced by temporal occlusion, the improvement is again more pronounced in association accuracy. Overall, the progressive gains across all three modules suggest that introducing complementary structural regularization into query interaction consistently benefits both detection and identity association.

\begin{table}[t]
\caption{\textbf{Ablation study on the spatial decay parameter $\sigma$ in Eq. (\ref{eq:geo}).}}
\label{exps:abla_temp}
\centering
\footnotesize
\begin{tabular}{c|ccc}
\toprule
$\sigma$     & HOTA & DetA & AssA \\ \midrule
0.25 & 66.1 & 79.8 & 55.0  \\
0.30 & 65.9 & 79.6 & 54.6  \\
\rowcolor{ourscolor} 0.35 & 66.9 & 79.2 & 56.7  \\
0.40 & 65.5 & 79.0 & 54.4  \\ \bottomrule
\end{tabular}
\vspace{-0.3cm}
\end{table}

\paragraph{Analysis of the Spatial Decay Parameter}
We further investigate the impact of the spatial decay parameter $\sigma$ in Eq.~(\ref{eq:geo}), which controls the spatial decay rate of the proximity modulation. As shown in Tab.~\ref{exps:abla_temp}, smaller values of $\sigma$ impose a more restrictive spatial bias. Although these settings slightly improve detection accuracy, they also suppress flexible association across larger inter-frame displacements, leading to inferior association performance. In contrast, excessively large $\sigma$ weakens the spatial regularization effect and introduces noisier long-range interactions during association, resulting in performance degradation across both DetA and AssA. Setting $\sigma$ to 0.35 achieves the best balance between spatial selectivity and interaction flexibility, yielding the highest overall tracking performance. Notably, this configuration is kept fixed across all evaluation benchmarks without dataset-specific tuning, suggesting that the proposed spatial modulation generalizes well across different tracking scenarios.

\begin{table}[t]
\caption{\textbf{Different Strategies for Historical Memory Update.}}
\label{exps:abla_hist_updata}
\centering
\footnotesize
\begin{tabular}{c|ccc}
\toprule 
History Update Weights & HOTA & DetA & AssA \\ \midrule
$w_\mathrm{base}$ & 65.2 & 79.7 & 53.5  \\ 
$\mathbf c$        & 65.2 & 79.9 & 53.3 \\ 
\rowcolor{ourscolor}   $w_\mathrm{base} \cdot \mathbf c$    &  66.9 & 79.2 & 56.7  \\ \bottomrule
\end{tabular}
\vspace{-4pt}
\end{table}

\paragraph{Analysis of Memory Update Strategy} 
Beyond regulating state integration under ambiguous interaction, the endogenous confidence vector $\mathbf c$ also governs the historical-memory update. Tab.~\ref{exps:abla_hist_updata} compares different memory update strategies while keeping the rest of the framework unchanged. Here, $w_\mathrm{base}$ is fixed to 0.3 across all experiments. Since the memory mechanism mainly affects temporal occlusion handling, its impact is primarily reflected in association accuracy. 
Using only the static momentum $w_\mathrm{base}$ yields limited association performance, suggesting that a fixed update rate cannot adapt to changes in interaction reliability. Conversely, using $\mathbf c$ alone can produce overly aggressive updates when reciprocal support is high, weakening stable long-term accumulation.
Combining the static base rate with the confidence modulation $w_\mathrm{base} \cdot \mathbf c$ achieves the best overall performance, reaching 56.7 AssA. This suggests that balancing stable long-term accumulation with adaptive reliability-aware updates is beneficial for maintaining temporal identity consistency.

\subsection{Mechanism Analysis}
In addition to standard ablations, we examine how the proposed modules shape query interaction and analyze the conditions under which the encoded structural regularities remain informative.

\begin{figure*}[pos=t]
    \centering
    \footnotesize
    \includegraphics[width=0.67\textwidth]{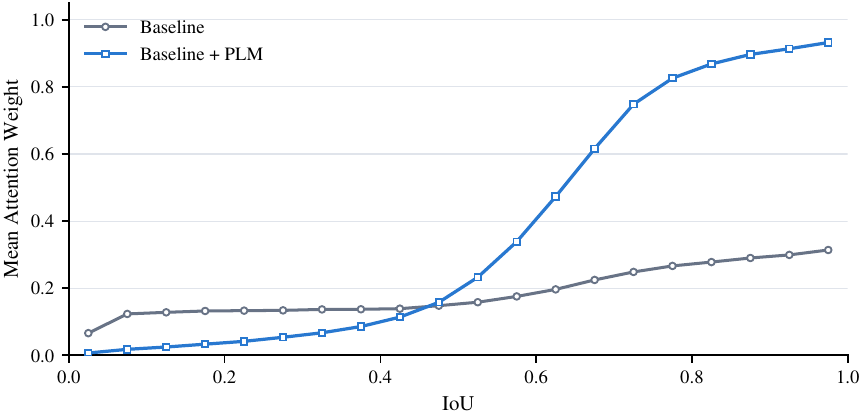}
    \caption{\textbf{Mean attention weights with respect to box IoU on the DanceTrack validation set.} The weights are derived from the track–detection interaction logits. Compared with the baseline, PLM produces clearer alignment between interaction strength and spatial overlap, indicating reduced responses to spatially implausible candidates during query interaction.
    }
    \label{fig:plm_analysis}
\end{figure*}

\paragraph{Analysis of PLM} To examine how PLM influences query interaction, we analyze the relationship between attention weights and box overlap, as shown in Fig.~\ref{fig:plm_analysis}. The baseline distributes attention across a broad range of overlap levels, including weakly overlapping candidates. After introducing PLM, attention weights become more concentrated around spatially compatible candidates and exhibit clearer alignment with box IoU. These observations suggest that PLM guides interaction toward spatially feasible regions and reduces responses to spatially implausible candidates.

\begin{table*}[t]
\centering
\footnotesize
\caption{\textbf{Analysis of QBAA on the DanceTrack validation set.} Lower normalized column entropy indicates less dispersed responses across competing track queries, a larger Top-1 margin reflects clearer separation between the dominant and competing responses, and a lower Multi-peak Ratio indicates fewer detection columns strongly shared by multiple tracks.}
\label{exps:tab_qbaa_analysis}
\begin{adjustbox}{max width=\textwidth}
\begin{tabular}{lccc}
\toprule
Setting 
& Column Entropy $\downarrow$
& Top-1 Margin $\uparrow$
& Multi-peak Ratio $\downarrow$ \\
\midrule
PLM                & 0.229 & 0.758 & 0.097 \\
\rowcolor{ourscolor} PLM + QBAA  & \textbf{0.112} & \textbf{0.842} & \textbf{0.085} \\
\bottomrule
\end{tabular}
\end{adjustbox}
\end{table*}

\paragraph{Analysis of QBAA} To examine whether QBAA alleviates ambiguous many-to-one competition during interaction, we analyze the column distributions of the aggregation weights used by the PLM and PLM+QBAA settings. Each column represents the responses of competing track queries to the same detection. For analysis, each column is normalized across track queries. We measure three statistics: normalized Column Entropy, which quantifies response dispersion and is normalized by the logarithm of the number of track queries; Top-1 Margin, which measures the difference between the largest and second-largest normalized responses; and Multi-peak Ratio, which measures the fraction of detection columns in which at least two responses reach 50\% of the maximum response in that column.

As shown in Tab.~\ref{exps:tab_qbaa_analysis}, introducing QBAA substantially reduces the column entropy while increasing the Top-1 margin. Meanwhile, the multi-peak ratio consistently decreases, indicating that fewer detections are simultaneously assigned strong responses by multiple competing tracks. These observations suggest that QBAA encourages clearer and less ambiguous interaction patterns during target association.

\begin{figure*}[pos=t]
    \centering
    \footnotesize
    \includegraphics[width=1.0\textwidth]{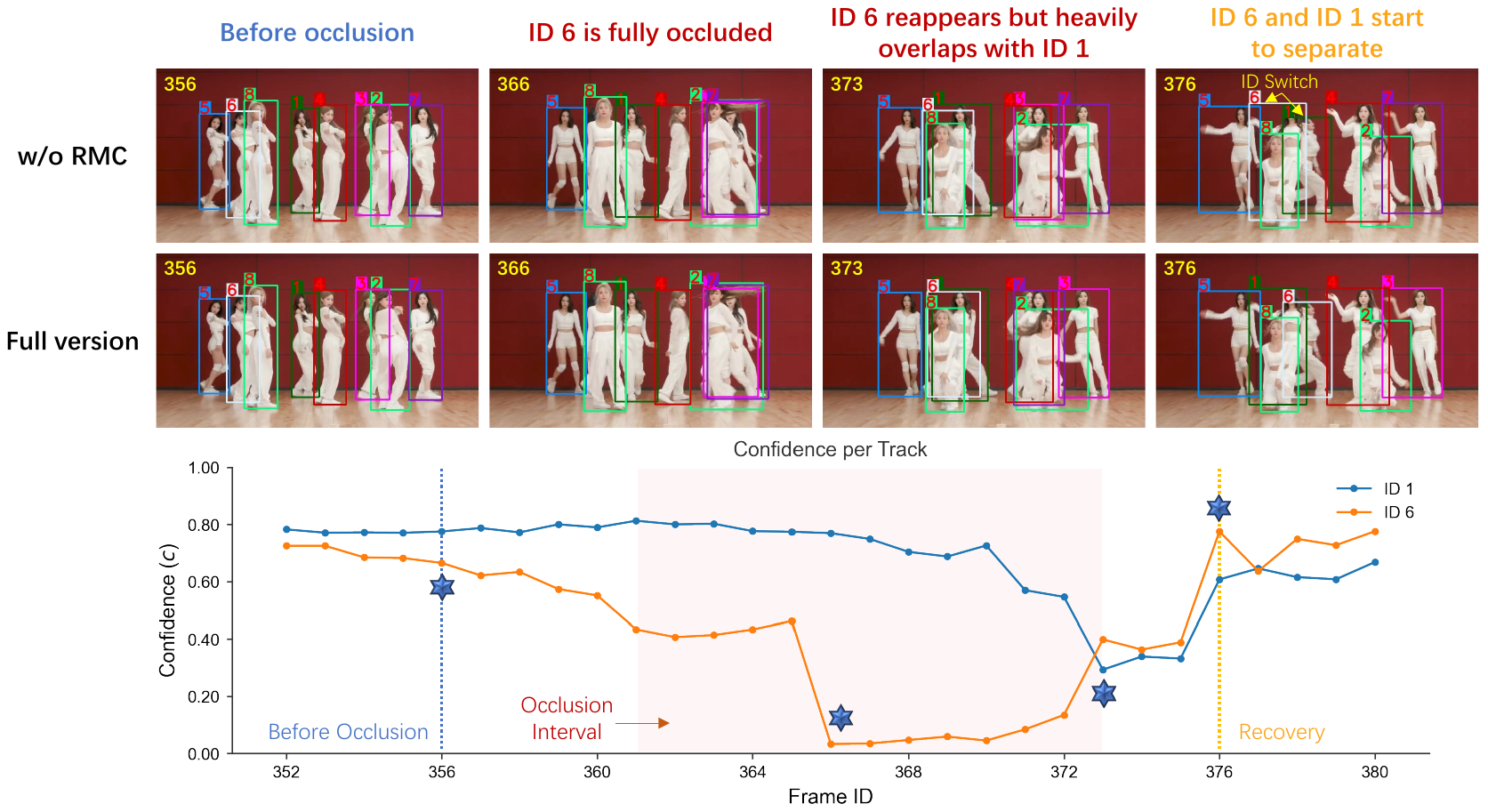}
    \caption{\textbf{Temporal analysis of RMC under severe occlusion on the DanceTrack validation set.} ID 6 becomes occluded around frame 361 and reappears near frame 373 with severe overlap with ID 1. The upper and lower rows show the results without RMC and with the full model, respectively. The curves report the interaction confidence $\mathbf c_t$ of the full model (Eq.~(\ref{eq:reliability})). The full model preserves identity continuity in this case, whereas removing RMC results in an ID switch.
    }
    \label{fig:vis_emm}
\end{figure*}

\paragraph{Analysis of RMC} To examine how RMC regulates state updating when track–detection interaction becomes unreliable, we visualize a representative occlusion–recovery case in Fig.~\ref{fig:vis_emm}. During frames 361–373, ID 6 becomes fully occluded and later reappears with severe overlap with ID 1, producing ambiguous interaction responses between tracks and detections. Without RMC, the tracker fails to recover the occluded identity and eventually produces an ID switch. With RMC, the interaction confidence derived from $\mathbf M_t$ decreases as reciprocal support weakens and gradually recovers as the two identities become distinguishable. RMC accordingly increases reliance on historical memory during ambiguous interaction, helping preserve identity continuity through the occlusion interval in this case.

\begin{table*}[pos=t]
\caption{\textbf{Tracking performance under temporal subsampling on the DanceTrack validation set.} GT IoU denotes the mean box overlap of the same identity in consecutive sampled frames. Subscripts indicate absolute improvements over the affinity-driven baseline.}
\label{exps:boundary_temporal}
\centering
\footnotesize
\setlength{\tabcolsep}{3.2pt}
\begin{tabular}{ccc|cccccc}
\toprule
\multicolumn{3}{c|}{Sampling condition}
& \multicolumn{3}{c}{Baseline}
& \multicolumn{3}{c}{PI-MOT} \\
\cmidrule(lr){1-3}
\cmidrule(lr){4-6}
\cmidrule(lr){7-9}
Stride & FPS & GT IoU
& HOTA & DetA & AssA
& HOTA & DetA & AssA \\
\midrule
1  & 20.00 & 0.91
& 59.3 & 76.2 & 46.4
& $66.9_{+7.6}$ & $79.2_{+3.0}$ & $56.7_{+10.3}$ \\
2  & 10.00 & 0.85
& 59.4 & 75.9 & 46.7
& $67.1_{+7.7}$ & $79.3_{+3.4}$ & $56.9_{+10.2}$ \\
4  &  5.00 & 0.76
& 60.6 & 74.2 & 49.7
& $66.2_{+5.6}$ & $77.4_{+3.2}$ & $56.9_{+7.2}$ \\
6  &  3.33 & 0.69
& 59.2 & 71.7 & 49.0
& $62.5_{+3.3}$ & $75.5_{+3.8}$ & $51.9_{+2.9}$ \\
8  &  2.50 & 0.64
& 57.1 & 69.7 & 46.9
& $59.8_{+2.7}$ & $72.2_{+2.5}$ & $49.6_{+2.7}$ \\
10 &  2.00 & 0.60
& 52.3 & 67.4 & 40.7
& $55.1_{+2.8}$ & $69.1_{+1.7}$ & $44.0_{+3.3}$ \\
\bottomrule
\end{tabular}
\vspace{-2pt}
\end{table*}

\paragraph{Boundary Analysis} PI-MOT encodes structural regularities to guide learnable query interaction, but their benefits may diminish under extreme tracking conditions. Inaccurate localization or abrupt displacement may weaken spatial discrimination or favor an incorrect candidate; when all candidates have zero overlap, the IoU-based bias provides no additional cue for distinguishing among them. Reciprocal support helps regulate competition but does not establish identity correctness. Incorrect correspondences may therefore receive strong interaction support and contaminate track memory. Persistent low confidence may also limit memory updates, leaving the stored state outdated.

To systematically assess the robustness of the proposed structural regularization, we progressively weaken temporal continuity of identities through frame subsampling. As shown in Tab.~\ref{exps:boundary_temporal}, we evaluate PI-MOT and the affinity-driven baseline on the DanceTrack validation set using sampling strides $s \in \{1,2,4,6,8,10\}$, corresponding to input frame rates of $20/s$ FPS. Starting from the first frame, we retain one frame every $s$ frames and consistently reindex the images and GT annotations. For each tracker, the checkpoint trained under the standard 20-FPS protocol is fixed across all sampling strides, without retraining. Tracking is rerun on each sampled sequence and evaluated against its corresponding GT. All inference hyperparameters, including track-lifecycle settings, remain unchanged.

\begin{figure*}[pos=t]
    \centering
    \footnotesize
    \setlength{\abovecaptionskip}{-0.3pt}
    \setlength{\belowcaptionskip}{-6pt}
    \includegraphics[width=1.0\textwidth]{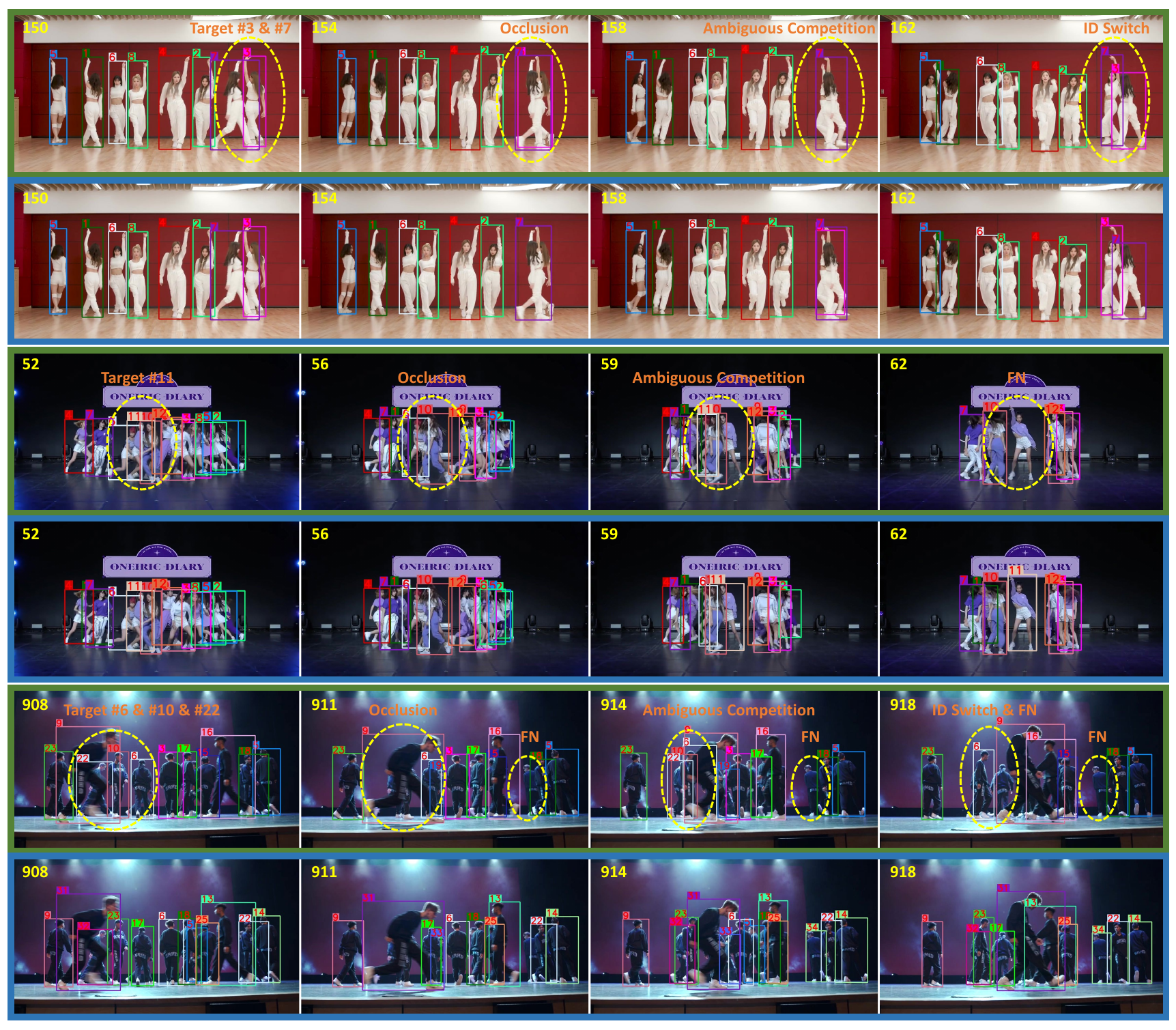}
    \caption{
    Qualitative analysis of representative challenging cases on DanceTrack. The first row of each case corresponds to the baseline, and the second row shows the results of PI-MOT.
    }
    \label{fig:vis_cases}
\end{figure*}

As the input frame rate decreases, the mean GT IoU falls from 0.91 to 0.60, indicating progressively weaker spatial continuity. PI-MOT maintains HOTA between 66.2 and 67.1 at 5–20 FPS, followed by a more pronounced decline under sparser observations. Its advantage over the baseline generally narrows: the HOTA and AssA gains decrease from 7.6 and 10.3 points at 20 FPS to 2.8 and 3.3 points at 2 FPS, respectively.
Despite this tenfold reduction in observation frequency, PI-MOT retains positive HOTA and AssA gains across all tested settings. The narrowing performance gap is consistent with the reduced informativeness of spatial continuity under increasingly sparse observations.
Together with the previous multi-benchmark comparisons, these findings support the practical value of encoding task-specific regularities as guidance for learnable interaction across diverse tracking conditions.

\subsection{Qualitative Analysis}

We further visualize several representative challenging cases on DanceTrack in Fig.~\ref{fig:vis_cases}. Under severe occlusion and dense interactions, missing or unreliable observations often cause the baseline tracker to gradually drift toward incorrect targets, eventually leading to ambiguous competition, ID switches, and failed target recovery during reappearance.

In contrast, PI-MOT exhibits more reliable identity continuity across representative challenging cases and remains more robust under long-term occlusion and heavy target overlap. These observations are consistent with the proposed structural regularization mechanisms for interaction modulation and reliability-aware temporal compensation.

\subsection{Discussion and Limitations}

The structural regularities encoded by PI-MOT arise from broadly applicable assumptions about object motion and identity association, but their effectiveness depends on the informativeness of the underlying cues. As illustrated by the temporal subsampling results in Tab.~\ref{exps:boundary_temporal}, the additional gains from structural regularization gradually diminish as inter-frame continuity weakens, yet remain positive even under substantially reduced observation frequencies.

PI-MOT is also affected by upstream detection errors. The proposed interaction regularization improves association using standard detector outputs without relying on detector-specific enhancement, but it is not designed to recover persistently missed detections or correct systematically biased boxes. In addition, the interaction confidence $\mathbf c$ summarizes instantaneous reciprocal support. It can regulate state updates under transient interaction ambiguity, but persistent, mutually supported incorrect correspondences may still produce high confidence because longer-term uncertainty is not explicitly modeled. Future work could extend confidence estimation with longer-range temporal consistency and improve robustness to persistent detection errors.

\section{Conclusion}

This work investigates how task-specific structural regularities can guide query interaction in tracking-by-query multi-object tracking. We reformulate query interaction as a structurally regularized information-routing process and introduce PI-MOT, which jointly regularizes the spatial selectivity, competitive structure, and temporal integration of observation information. This formulation preserves soft and differentiable interaction while promoting more reliable identity-oriented information routing.

Experiments on DanceTrack, SportsMOT, and MOT20 demonstrate strong tracking performance. Further analyses show improved data efficiency and faster convergence, while computational comparisons indicate modest overhead. Mechanism analyses confirm that the proposed components reshape query interaction in their intended directions. Controlled temporal-subsampling evaluation characterizes the operating boundary of the framework: PI-MOT remains advantageous across a tenfold reduction in observation frequency, although its additional benefit gradually narrows when inter-frame continuity is severely weakened. Overall, these findings demonstrate the value of integrating task-specific structural regularities into data-driven interaction learning, providing a practical direction for more robust and efficient end-to-end tracking.

\section*{Acknowledgements}
We are grateful for the support of the National Natural Science Foundation of China (No. 62271143), the Frontier Technologies R\&D Program of Jiangsu (No.BF2024060), and the Big Data Computing Center of Southeast University.

\section*{Declaration of Generative AI}
During the preparation of this work, the authors used ChatGPT to refine the academic phrasing, improve the overall grammatical structure of the manuscript, and ensure correct spelling and formatting. After using this tool, the authors thoroughly reviewed and edited the content as needed and take full responsibility for the ultimate content of the publication.

\printcredits

%\bibliographystyle{cas-model2-names}
\bibliographystyle{unsrtnat}

\bibliography{cas-refs}



\end{document}